# A Comparative Study of U-Net Architectures for Change Detection in Satellite Images

*Yaxita Amin[1*], Naimisha S Trivedi[2†] and Rashmi Bhattad[3†]*

[1] Department of Information Technology, Gujarat Technological University, Ahmedabad, India

[2] Department of Information Technology, Gujarat Technological University, Ahmedabad, India

[3] Gujarat Technological University, Ahmedabad, India

*[*]yaxita2003@gmail.com*

**Keywords**: Remote sensing, Change detection, U-Net architecture

## Abstract

Remote sensing change detection is essential for monitoring the everchanging landscapes of the Earth. The U-Net architecture has gained popularity for its capability to capture spatial information and perform pixel-wise classification. However, their application in the Remote sensing field remains largely unexplored. Therefore, this paper fill the gap by conducting a comprehensive analysis of 34 papers. This study conducts a comparison and analysis of 18 different U-Net variations, assessing their potential for detecting changes in remote sensing. We evaluate both benefits along with drawbacks of each variation within the framework of this particular application. We emphasize variations that are explicitly built for change detection, such as Siamese Swin-U-Net, which utilizes a Siamese architecture. The analysis highlights the significance of aspects such as managing data from different time periods and collecting relationships over a long distance to enhance the precision of change detection. This study provides valuable insights for researchers and practitioners that choose U-Net versions for remote sensing change detection tasks.

## 1 Introduction

Earth Observations provides globally consistent, repetitive measurements of earth surface conditions relevant to land cover monitoring [1]. Remote sensing is often viewed as an aid to landscape change detection and land use/cover classification [2]. This discovers changing landscapes and stable regions in the long run, it contributes to the preservation and improvement of human well-being, therefore remote sensing analysis is essential.

Change detection is the process of identifying differences in the state of an object or phenomenon by observing it at different times [2]. Several approaches have been devised for change detection: Image differencing, Vegetation Indices (e.g., NDVI, SAVI), Principal Component Analysis (PCA), Change Vector Analysis (CVA), and Deep Learning-based Techniques. With rapid development, a huge amount of spectral–spatial-temporal data raises demand for the study of deep learning approaches for effective analysis. CNN, a remarkable architecture extracts features from complex spectral–spatial-temporal information within data. We investigate various CNN architectures and their capabilities[3], [4], [5], [6], [7].

FCN unlocked the potential of image segmentation. One of the most powerful segmentation architectures nowadays is U-Net[8]. U-Net handles complex and high spectral–spatial-temporal resolution data while retaining spatial information. By analyzing high-resolution data, it discovers changes in landscapes over time with high accuracy. Pixel-wise segmentation allows us to understand more about changes than traditional methods. As development is ongoing on U net, several variants have been made to enhance the performance of the model.

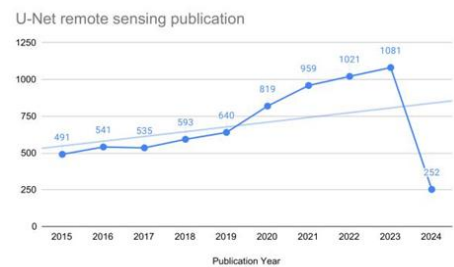

*Fig. 1 U-Net variants Trend in last past years*

As shown in Figure 1, there has been a substantial increase in the publication of U-Net modifications in the remote sensing field that shows growing interest in leveraging its effectiveness. Motivated by baseline U-Net, this paper, therefore, answers these questions:
1)How effective are existing U-Net modifications?
2)How it will respond when it is applied to remote sensing change detection, that originally developed for medical segmentation tasks?
3)Can advanced U-Net variants specifically designed for the RS field outperform base architecture?

By comparing and exploring different U-Net variants, this comparative analysis aims to contribute valuable knowledge in the RS field.

## 2. Background

Remote sensing change detection provides a rich set of functionalities as a development that is constantly evolving. A wide spectrum of techniques is available, however, deep learning methods are still better in many ways due to their effective learning capability and adapt to diverse types of data [9].Following the development of U-Net: Convolutional Networks for Biomedical Image Segmentation, U-Net architecture paved the way for unseen possibilities for remote sensing analysis.

U-Net was built upon a more elegant architecture, the so-called "fully convolutional network" [8]. Due to its capability to capture high-level semantic information as well as low-level spatial data it is originally proposed by [8]. Below shown table outlines important components of U-Net.

| Component | Description |
| --- | --- |
| Encoder | An important network for feature extraction. Uses convolutional layers to generate feature maps at different resolutions. |
| Decoder | Upsamples the extracted features and combines them with skip connections from the encoder for precise localization. |
| Skip Connections | Direct connections between corresponding levels in the encoder and decoder. |

*Table 1 Component Descriptions*

U-Net architecture Accurate segmentation can be challenging, to address these challenges U-Net tackles limitations through its key functionalities. U-Net addresses these challenges: 1) Limited dataset: Due to the efficient use of feature channels and skip connections, U-Net can potentially achieve good performance even with small datasets. This model therefore uses data augmentation, it employs elastic deformation. 2) Touching Object: U-Net excels in this difficult task of accurately segmenting objects that are overlapping or touching by incorporating a weighted loss function. 3) Loss of spatial information: Classical CNN with a pooling layer suffers from loss of information in the downsampling part, mirroring mitigates to help this issue.

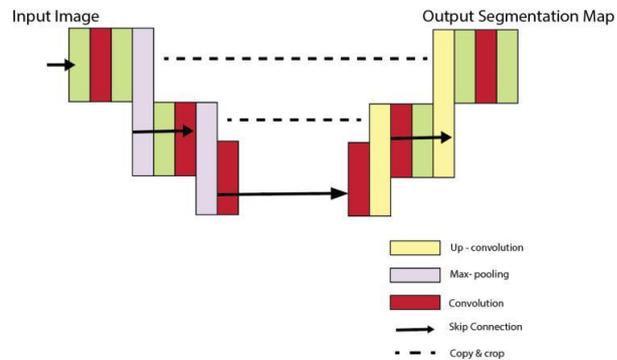

*Fig. 2 U-Net architecture*

The Key Innovation of this model in the upsampling part is a large number of feature channels that allow networks to propagate context information to higher-resolution images. As shown in Figure 2, the left side is the contracting path that follows the typical architecture of a convolutional network. The pooling layer performs downsampling that reduces the spatial resolution of feature maps while increasing the number of feature channels by zooming in on important details within smaller regions, this helps the network to capture high-level features that are important for segmentation. Unlike traditional CNNs, U-Net recovers lost spatial resolution during upsampling (Right side path). Accurate feature extraction is an essential step during segmentation. Furthermore, skip connections can be viewed as shortcut paths because they bridge the gap between the contracting path and the expanding path. Skip connections concatenate feature maps from the encoder with the corresponding levels in the decoder. It helps to preserve crucial information lost during downsampling.

*2.1 Limitations of Standard U-Net*

Firstly, the receptive field of U-Net is limited, as a result, it struggles to capture long-range dependencies between features in remote sensing images. This can be problematic in RS because CD tasks involve analyzing information from larger spatial contexts across past and present images. Secondly, because of the inherent limitation of convolution operations, U-Net compromised in learning global context and long-range spatial relations. CD's crucial task is to analyze overall scenes and relationships between different areas so this limitation could lead to complications. Additionally, The skip connection is too sloppy, the features of the two convolution layers being connected have large semantic differences, which leads to increased learning difficulty of the network [10]. Lastly, U-Net has drawbacks like large model size, and the use of fixed size kernels that can only be used if the size of salient regions in the images is the same.

## 3 U-Net variants

U-Net is a capable technique for various image segmentation tasks, however, for change detection limitations arise due to inherent characteristics of remote sensing imagery. To address the above-mentioned limitations, researchers started to develop several variants as the need of different fields. In the early stage of development, pioneering modifications like ResNet , SegNet [11] and DenseNet [12] paved the way for U-Net's flourishing development. In this paper, we pro-posed recent developments in U-Net which is used in remote sensing change detection for robust performance of the model.

*3.1 Foundation of U-Net variants*

These architectures provide an essential foundation in U-Net variations: Stack U-Net [13] incorporates multi-resolution feature extraction by merging high-level resolution details with low-level contextual information across multiple stacks.

Compared to CNNs, it extracts fewer focused feature maps. Furthermore, a context aggregation module assists with a fusion of context signals. Integrates attention mechanism during concatenation aids effective combination of features.

U-Net: Transformer [14] addresses memory intensiveness by selecting strides convolutional layers instead of pooling layers. It utilized two-level depth U-Net encoder-decoder architecture, facilitating the recovery of spatial info from rich semantic features. Two types of attention modules are mentioned in this paper [14] : Multi-Head Self-Attention (MHSA) located at the end of the encoder, focuses on capturing long-range dependencies, and Multi-Head Cross-Attention (MHCA) aims to improve the efficiency of U-Net by integrating semantic richness from the high-level feature with high resolution from skip connection while filtering out noisy area.

Dense U-Net [12] integrates dense blocks for feature extraction in dense sampling by replacing pooling layers. Dense blocks allow for features to directly connect to each layer to all preceding layers within the block, which enhances the information flow in the network [15]. This U-Net strengthens feature propagation to learn the more complex relationships between features.

Recurrent Residual U-Net [16] incorporates recurrent convolutional layers within the encoder path and residual connection to improve performance. Recurrent Connections are adjusted to handle the process of a sequence of temporal images to learn features over time. Residual Connections integrates within the encoder and decoder path to extract complex relationships of features.

Attention U-Net [17] integrated within standard U-Net. Attention gates learn to focus on relevant image regions automatically during training without additional suspension. Dense label predictions enable pixel-wise segmentation of the entire image. AGs placed before the concatenation operation in skip connection of U-Net architecture. This ensures that only task relevant features are propagated through the network.

U-Net++ [18] improves information flow through the network by integrating dense skip pathways. Where each encoder passes data to all decoder layers, it preserves spatial details. U-Net++ offers multi-level feature maps and deep supervision to improve training.

U SE-Net [19], [20], [21] proposed a network by adding SE blocks after the encoder and decoder. It affects low-level features in U-Net architecture and increases overall performance. SE blocks introduce a channel attention module that learns importance of different feature channels within convolutional layers.

Deep residual U-Net [22] is utilized by substituting plain unit with residual unit. Residual units going deeper would improve the performance of multilayer neural networks. It will ease training the network. Inspired by ResNet[23] , it allows the network to learn from original inputs directly even in deeper layers.

Multi-class U-Net [24] is not only identifying but also differentiating between multiple types. The output segmentation layer is expanded from 1 to N feature maps to enable multi-class segmentation. Batch normalization used in all layers to improve performance. The number of feature maps has been halved compared to the original architecture to speed up the performance.

3D U-Net [25] specifically designed for volumetric segmentation tasks - sequences of images captured at different time points. This not only learns spatial features within the image but also the temporal changes occurring between them. It can learn effective segmentation models even with sparse annotations, significantly reducing time and effort.

*3.2 A detailed categorization Of Advanced U-Net Variants*

RS CD aims to analyze areas where the landscapes change over time has undergone modifications between two or more temporal images captured. This section categorizes them based on their key modifications that contribute to improved change detection performance.

As shown in Figure 3, illustrates the compendium of modifications, showcasing a range of potential alternations to base architecture. [14], [16] proposed architecture with attention and transformer module (denoted by the A and T blocks in Figure 3) potentially capture important features from input data. Notably, it can be either placed in the encoder, decoder or skip connection part.

Another way of modification is in/on base architecture that parallel U-Net (label as the "P" block)[22]. Here in Figure 3, H represents a hierarchical module that incorporates multiple U-Net in a nested manner [10]. Another way of modification, placing an additional block (eg SE [18] Dense block[15] ) after each encoder and decoder to customize the network's feature extraction.

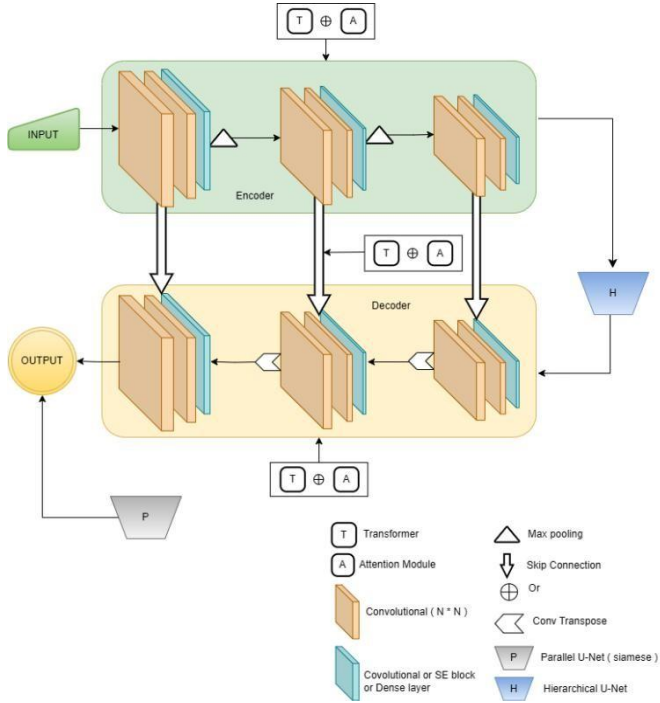

*Fig. 3 Visualizing Modifications to the U-Net Architecture*

| Model | U-Net-Modification | Transfer Learning | Dense Layers in Encoder/Decoder |
|---|---|---|---|
| Siamese Swin-U-Net | None | No | Yes (Encoder & Decoder) |
| STCD-EffV2T U-Net | None | Yes (EffV2-T) | Yes (Decoder) |
| T-U-Net | Triple branch architecture | No | Yes (Encoder & Decoder) |
| Ensemble Net-ResNet | U-Net within ResNet , Ensemble of U-Net &ResNet | Yes (U-Net and ResNet) | Yes (U-Net and ResNet) |
| Optimised U-Net | Modifications to skip connections | No | Yes (Encoder & Decoder) |
| Bilateral Attention U-Net | Attention mechanism in encoder | No | Yes (Encoder & Decoder) |
| NDR-U-Net | Channel-wise non-local attention | No | Yes (Decoder) |
| HARNU-Net | Hierarchical Attention Mechanism for feature extraction | No | Yes (Encoder & Decoder) |

*Table 2 Modification Strategies for U-Net Variants in Remote Sensing Change Detection*

Explanations of Table 2 as follows: 1. Attention-based U-Net Variants: Attention mechanism allows model to "pay attention" to certain parts of data and give them more weight during training, indicating their relevance while improving accuracy and computational cost. It calculates similarity scores between pixels in the concatenated feature maps. Similarity score measured by two methods spatial attention [25] it focuses on the relationship between pixels, captures local dependencies and channel attention, it gives importance to channels within feature maps. Based on the calculated similarity score, the attention module generates weight map. Siamese Swin [26]. It replaces standard convolutional layers in the encoder path with a Swin Transformer block. Swin Transformer blocks are specifically designed to capture long-range dependencies due to their shifted window-based self-attention mechanism. Because of this block, Siamese Swin-U-Net can effectively identify changes in remote sensing images by focusing on features with relevant spatial relationships across the image. Bilateral Attention U-Net [27] : This variant integrates a bilateral attention mechanism within the skip connections. Skip connections in U-Nets typically concatenate features from the encoder and decoder paths at different resolutions. The bilateral attention mechanism refines this process by focusing on informative features from both paths. This allows the network to prioritize features that are most relevant for change detection, leading to more accurate results.

2. Transfer Learning U-Net Variants: Transfer learning is powerful technique for leveraging pre-trained models in remote sensing tasks. By utilizing knowledge from large dataset, U-Net variants can achieve high performance with less training data. In transfer learning, crucial step is to choose right pre-trained model suitable to remote sensing task. The pre-trained model must have a strong feature extraction capability. These variants leverage pre-trained models on large datasets to improve performance on smaller remote sensing datasets. Remote sensing datasets are often limited in size, which can hinder the effectiveness of deep learning models. While in training, weights are freezed by encoder to ensure model as fixed feature extractor. It allows these variants to take advantage of pre-trained feature representations from a different task or dataset.

STCD-EffV2T U-Net [28]: It utilizes EfficientNetV2 (potentially pre-trained) as the backbone in the encoder path. EfficientNetV2 is a CNN architecture pre-trained on a massive image dataset (ImageNet). By using this pre-trained model, STCD-EffV2T U-Net can leverage the learned feature representations in RS CD, even with limited training data. This can definitely improve the accuracy and efficiency of the model compared to training from scratch.

3. U-Nets with Specific Encoder Modifications: In encoder path, baseline U-Net struggles to capture complex feature for task like VHR/HR images in remote sensing. These variants introduce modifications within the encoder path to enhance feature extraction for change detection tasks. T-U-Net [29] This variant integrates a transformer module in the decoder path. While not strictly an encoder modification, it can be considered a specific change that impacts feature processing. Transformer modules are known for their ability to capture long-range dependencies and global context. By incorporating a transformer module in the decoder, T-U-Net can focus on informative features across the entire image, enhancing global context learning during upsampling. This can lead to improved change detection performance, especially for identifying large-scale changes in the landscape.

4. U-Nets with Specific Skip Connection Modifications: it focus on refining the information flow within skip connections to improve change detection accuracy. For example, Optimised U-Net [29] ,introduces bottleneck structures within skip connections. Skip connections can struggle with dimensionality issues when combining features from different resolutions. Optimised U-Net addresses this by introducing bottleneck structures that reduce the dimensionality of features before concatenation. This improves the information flow within skip connections. Several optimization techniques enhance the performance of model by optimizing different aspects of base architecture [30], [31].

Bilateral Attention U-Net [27]: As discussed earlier, Bilateral Attention U-Net also falls under this category. Its bilateral attention mechanism within skip connections focuses on informative features from both the encoder and decoder, leading to improved information flow for change detection.

5. Other modifications: For boosting accuracy of baseline U-Net, many researchers are modifying on and upon baseline U-Net architecture. With 3 main components of U-Net, they are trying to modify and manipulate among those parts.

NDR-U-Net [32]: it likely introduces modifications in the encoder path. It might utilize multi-dilated convolutional layers to capture features at different scales. Multi-dilated convolutions can be beneficial for capturing long-range dependencies in remote sensing images.

HARNU-Net [10]: this variant introduces modifications in both the encoder and skip connections. Encoder Modifications: this model employs cascaded residual dense blocks in the encoder. These blocks connect multiple convolutional layers with dense connections (all layers connected to each other) and residual connections (adding the input to the output). This structure can improve feature extraction by allowing better information flow and potentially mitigating vanishing gradients in deeper networks. Skip Connection Modifications: it also utilizes a channel and spatial attention mechanism within skip connections. This mechanism focuses on informative features across channels (feature maps) and spatial locations within the feature maps. This improves the flow of relevant features through skip connections.

*3.3 Challenges in Remote Sensing Change Detection*

Detecting changes in remote sensing images is a complex challenge due to various factors, including variations in image quality, noise, registration errors, illumination changes, complex landscapes, and spatial heterogeneity[33]. Change detection poses some significant challenges as paired images are often captured under varying conditions, such as different angles, illuminations, and even during different seasons, resulting in diverse and unknown changes in a scene and even it captured by various sensors. These challenges encompass a broad range of issues, including 1) variations in image quality arising from differences in spatial-spectral-temporal resolutions; 2) the presence of diverse types of noise and artifacts; 3) errors during image registration; 4) difficulties in handling illumination, shadows, and changes in viewing angle; 5) complex and dynamic landscapes; 6) scale and spatial heterogeneity of the landscape[33]; 7) filtering out the relevant data [10]; 8) high computational demands of complex models; 9) a lack of interpretability in their predictions [34].

# 4 Solutions to Various Problems

This section will explore how advanced U-Net variants address the challenges of U-Net and RS CD to achieve improved performance.

| Architecture Name | Dataset Name (Source) | Sensor | Resolution (m) | Location | Satellite Type |
|---|---|---|---|---|---|
| Siamese Swin-U-Net | CDD (Google Earth Engine) | - | 0.03 - 1 | - | - |
| STCD-EffV2T U-Net | OSCD + Sentinel-2 (Iran) | Sentinel-2 | 0.1 and 0.6 | Iran | Sentinel-2 |
| T-U-Net | LEVIR-CD, WHU-CD, DSIIN-CD | - | 0.5, 0.2 | 0.3, Texas, USA | - |
| Ensemble U-Net-ResNet | HRS (WorldView-2) | WorldView-2 | 0.5 | Beijing, China | WorldView-2 |

| Architecture Name | Dataset Name (Source) | Sensor | Resolution (m) | Location | Satellite Type |
|---|---|---|---|---|---|
| Optimised U-Net | Vaihingen (LIDAR) | LIDAR | 0.2, 0.25, 0.5 | Vaihingen (Germany) | LIDAR |
| Bilateral Attention U-Net | Gaofen-2 (Image source) | Gaofen-2 | 0.8 | Seoul (Korea) | Gaofen-2 |
| NDR-U-Net | GF-2 | GF-2 | 0.8 | Europe region | GF-2 |
| HARNU-Net | GF-2 (China) | GF-2 | 0.8 | New Zealand, Texas (USA) | GF-2 |

*Table 3 Comparison of different datasets of U-Net RS variants*

Table 3. showcases the diversity of datasets employed for evaluating U-Net variants. We observe a range of resolutions (0.03m - 3m), different sensors (optical, LiDAR), and diverse landscapes ( USA, Iran). This provides valuable insights into the adapt-ability of U-Net variants across various remote sensing scenarios. Variants that rely more on learning complex feature maps from a wide number of classes might underperform datasets with restricted sets. Multi-source data fusion like combining optical and LiDAR can provide richer information. In this case, optical imagery offers high spatial resolution for capturing efficient land cover features, while LiDAR data illustrates precise elevation information.

*4.1 Limitations Of U-Net Address by Variants*

- Limited dataset problem where in baseline U-Net standard convolutional layers limiting network's ability to capture long-distance relationships in RS images. Where Siamese Swin-U-Net integrated Swim Transformer architecture in the Siamese network. The transformer is capable of capturing long-range dependencies and learning relationships between features. STCD-EffV2T U-Net [28] offers a good balance between accuracy and computational efficiency.

- Global context issue arise where U-Net struggles to learn global information effectively. T-U-Net [35] incorporates a transformer within the decoder path. It allows the network to focus on informative features relevant to change detection across the entire image. Ensemble U-Net-ResNet combines prediction from multiple U-Net architectures with ResNet backbones.

- "Sloppy" skip connection can lead to learn difficulties. Optimised U-Net [29] introduces a bottleneck structure within skip connections, and helps to reduce dimensionality of features being concatenated, potentially mitigating issues. Bilateral Attention U-Net [27] integrates bilateral attention mechanisms within skip connections. It allows the network to focus on informative features. Where HARNU-Net [10] addressed limitations like limited receptive field and skip connection issues.

| Architecture Name | Evaluation Metric | Performance Score | Future Scope |
|---|---|---|---|
| Siamese Swin-U-Net | F1-Score | 94.67 | Investigation on Siamese architecture for multi-temporal LULC change detection |
| STCD-EffV2T U-Net | Mean IoU | 0.87 | Explore pre-trained EfficientNetV2 for various land cover classification tasks |
| T-U-Net | Pixel Accuracy | Higher accuracy | Compare the effectiveness of triplet loss for LULC classification tasks and specialized loss functions |
| Ensemble U-Net-ResNet | U-Net-Overall Accuracy | 89.20% | Explore segmentation techniques from ResNet for improved delineation of LULC classes |
| Optimised U-Net | Overall Accuracy | 92.10% | Investigate the transferability of optimizations to various land cover |
| Bilateral Attention U-Net | Mean IoU | 0.845 | Accurate and explore DAG for land cover |
| NDR-U-Net | Mean IoU | 0.92 | Investigate the effectiveness of nested dense residual blocks |

| Architecture Name | Evaluation Metric | Performance Score | Future Scope |
|---|---|---|---|
| HARNU-Net | Mean IoU | 0.88 | Explore combining hierarchical attention with land cover-specific features |

*Table 4 Comparison of Evaluation matrices of U-Net Variants*

Table 4 highlights the performance of each variant, as shown in the table dominant metrics Mean IoU (Intersection over Union) for segmentation task and F1-Score for classification tasks. From the table, noticeably NDR-U-Net achieves the highest Mean IoU (0.92) indicating excellent segmentation accuracy. Siamese Swim-U-Net illustrates a string F1-Score (94.67) for LULC change detection. Optimized U-Net shows promising overall accuracy (92.10 %) for land cover classification. These are top contenders from these mentioned recently developed variant models.

*4.2 Limitations Of RS CD Address by Variants and various techniques*

Variations in image quality: U-Net variants with deeper architectures, for instance, Stacked U-Net [13] and Dense U-Net [12] can learn complex features regardless of variations in image quality. Utilizing data augmentation alongside U-Net variants can artificially increase the amount and variety of training data, mitigating the effects of limited and potentially noisy data. Addressing Spectral and Illumination Variations: Spectral Variability: U-Net variants incorporate techniques like normalization (e.g., min-max normalization) that can enhance.

feature extraction by normalizing the spectral information in remote sensing images. Attention module, which focuses on important information within the normalized spectral bands. Illumination Changes: U-Net with Transformer and U-Net bilateral Attention can potentially learn illumination variations by capturing long-range dependencies. Addressing Landscape Complexity: Complex and Dynamic Landscapes: Deeper architectures like Stacked U-Net, U-Net++ [18]and HARNU-Net can potentially discern intricate features from a variety of land cover types within complex landscapes. Using convolutional layers with larger kernels in U-Net ResNet architecture can enhance performance.

## 5 Discussion & Conclusion

U-Net core ten variants are working effectively in medical segmentation tasks. Motivated by these variants, shifting focus towards remote sensing we study 8 variants. The research highlights the success of Optimized U-Net and NDR U-Net. As for Optimized U-Net, to learn from the variant is to emphasize specific parameters for specific applications, ensuring the model is well-suited for data. NDR U-Net is basically pro-posed for water body segmentations but can be applicable for LULC change detection applications as it has a variety of benefits because nested layers have dense blocks that identify complex features with ease.

A critical consideration during variant selection is the balance between computational efficiency and model complexity. Deeper architectures with complex attention mechanisms can capture important features and usually demand high computational power. Selecting the right U-Net variant necessitates understanding the trade-off between accuracy and efficiency based on the specific characteristics of the Remote Sensing change detection task. This research opens the door to a number of possibilities of investigating the mentioned variants in remote sensing applications